\documentclass{article}

\usepackage{microtype}
\usepackage{graphicx}
\usepackage{booktabs} %
\usepackage{epsfig}
\usepackage{endnotes}
\usepackage{subcaption}
\usepackage{xspace}
\usepackage{float}
\usepackage{algorithm}
\usepackage{algorithmic}
\usepackage{amsmath}
\usepackage{ifdraft}
\usepackage{adjustbox}
\usepackage{stfloats}

\PassOptionsToPackage{hyphens}{url}
\usepackage[final]{hyperref}

\usepackage[accepted]{icml2025}

\usepackage{amsmath}
\usepackage{amssymb}
\usepackage{mathtools}
\usepackage{amsthm}
\usepackage{subcaption}

\usepackage[capitalize,noabbrev]{cleveref}

\crefname{section}{Sec.}{Secs.}
\Crefname{section}{Sec.}{Secs.}
\crefname{figure}{Fig.}{Figs.}
\Crefname{figure}{Fig.}{Figs.}
\crefname{table}{Tab.}{Tabs.}
\Crefname{table}{Tab.}{Tabs.}
\crefname{appendix}{App.}{Apps.}
\Crefname{appendix}{App.}{Apps.}
\crefname{equation}{Eq.}{Eqs.}
\Crefname{equation}{Eq.}{Eqs.}
\crefname{algorithm}{Alg.}{Algs.}
\Crefname{algorithm}{Alg.}{Algs.}

\theoremstyle{plain}

\theoremstyle{definition}

\theoremstyle{remark}

\usepackage[textsize=tiny]{todonotes}

\newcommand{\sys}{Tactic\xspace}
\newcommand{\quest}{Quest\xspace}

\newcommand{\pyramid}{PyramidKV\xspace}
\newcommand{\adakv}{Ada-SnapKV\xspace}

\newcommand{\as}{attention score}

\newcommand{\AS}{Attention Score}
\newcommand{\fig}[1]{Fig.~\ref{#1}}

\newcommand{\tab}[1]{Tab.~\ref{#1}}
\newcommand{\df}{distribution fitting\xspace}

\newcommand{\DF}{Distribution Fitting\xspace}
\newcommand{\attnspeedup}{$7.29\times$\xspace}
\newcommand{\etoespeedup}{$1.58\times$\xspace}

\icmltitlerunning{\sys: Adaptive Sparse Attention with Clustering and Distribution Fitting for Long-Context LLMs}

\begin{document}

\twocolumn[
\icmltitle{\sys: Adaptive Sparse Attention with Clustering and Distribution Fitting for Long-Context LLMs}

\icmlsetsymbol{equal}{*}

\begin{icmlauthorlist}
\icmlauthor{Kan Zhu}{equal,uw}
\icmlauthor{Tian Tang}{equal,uw,thu}
\icmlauthor{Qinyu Xu}{equal,uw,thu}
\icmlauthor{Yile Gu}{uw}
\icmlauthor{Zhichen Zeng}{uw}
\icmlauthor{Rohan Kadekodi}{uw}
\icmlauthor{Liangyu Zhao}{uw}
\icmlauthor{Ang Li}{uw}
\icmlauthor{Arvind Krishnamurthy}{uw}
\icmlauthor{Baris Kasikci}{uw}
\end{icmlauthorlist}

\icmlaffiliation{thu}{Tsinghua University}
\icmlaffiliation{uw}{University of Washington}

\icmlcorrespondingauthor{Baris Kasikci}{baris@cs.washington.edu}

\icmlkeywords{Large Language Model, Sparse Attention, Inference, KV Cache}

\vskip 0.3in
]

\printAffiliationsAndNotice{\icmlEqualContribution} %

\begin{abstract}

Long-context models are essential for many applications but face inefficiencies in loading large KV caches during decoding. Prior methods enforce fixed token budgets for sparse attention, assuming a set number of tokens can approximate full attention. However, these methods overlook variations in the importance of attention across heads, layers, and contexts.

To address these limitations, we propose \sys, a sparsity-adaptive and calibration-free sparse attention mechanism that dynamically selects tokens based on their cumulative attention scores rather than a fixed token budget. By setting a target fraction of total attention scores, \sys ensures that token selection naturally adapts to variations in attention sparsity. To efficiently approximate this selection, \sys leverages clustering-based sorting and distribution fitting, allowing it to accurately estimate token importance with minimal computational overhead.

We show that \sys outperforms existing sparse attention algorithms, achieving superior accuracy and up to \attnspeedup decode attention speedup. This improvement translates to an overall \etoespeedup end-to-end inference speedup, making \sys a practical and effective solution for long-context LLM inference in accuracy-sensitive applications.

\end{abstract}

\section{Introduction}
\label{sec:intro}

Large language models (LLMs) power a wide range of applications, from conversational assistants to document analysis systems and search engines. The demand for multi-turn interactions and long-document processing has driven an expansion of context length, growing from thousands to as many as one million tokens~\cite{liu2024scalinglawsropebasedextrapolation}.

However, supporting long contexts in LLM inference presents significant challenges, primarily due to the growing memory footprint of the Key-Value (KV) cache~\cite{tang2024questqueryawaresparsityefficient}. The memory requirements of the KV cache scale proportionally with the context length, therefore, it can quickly become a bottleneck despite optimizations such as Grouped-Query Attention (GQA)~\cite{ainslie2023gqa}. Furthermore, the need to repeatedly load the KV cache for every generated token becomes a bottleneck. For instance, loading the large KV cache can account for over 50\% of the total latency during auto-regressive decoding, significantly impeding the efficiency of large-scale serving systems.~\cite{tang2024questqueryawaresparsityefficient}

To mitigate the high cost of KV-cache loading, recent methods approximate full attention by selecting a subset of stored Key and Value vectors, corresponding to a subset of tokens, within a fixed token budget~\cite{liu2024clusterkvmanipulatingllmkv,tang2024questqueryawaresparsityefficient,zhang2023h2o,xiao2023streamingllm}. These approaches exploit the natural sparsity of attention, where only a small fraction of tokens significantly influence the output due to the softmax operation. By leveraging this sparsity, they aim to reduce the overhead of loading the KV-cache without sacrificing model accuracy.

\begin{figure}[t]
    \centering
    \includegraphics[draft=false, width=\columnwidth]{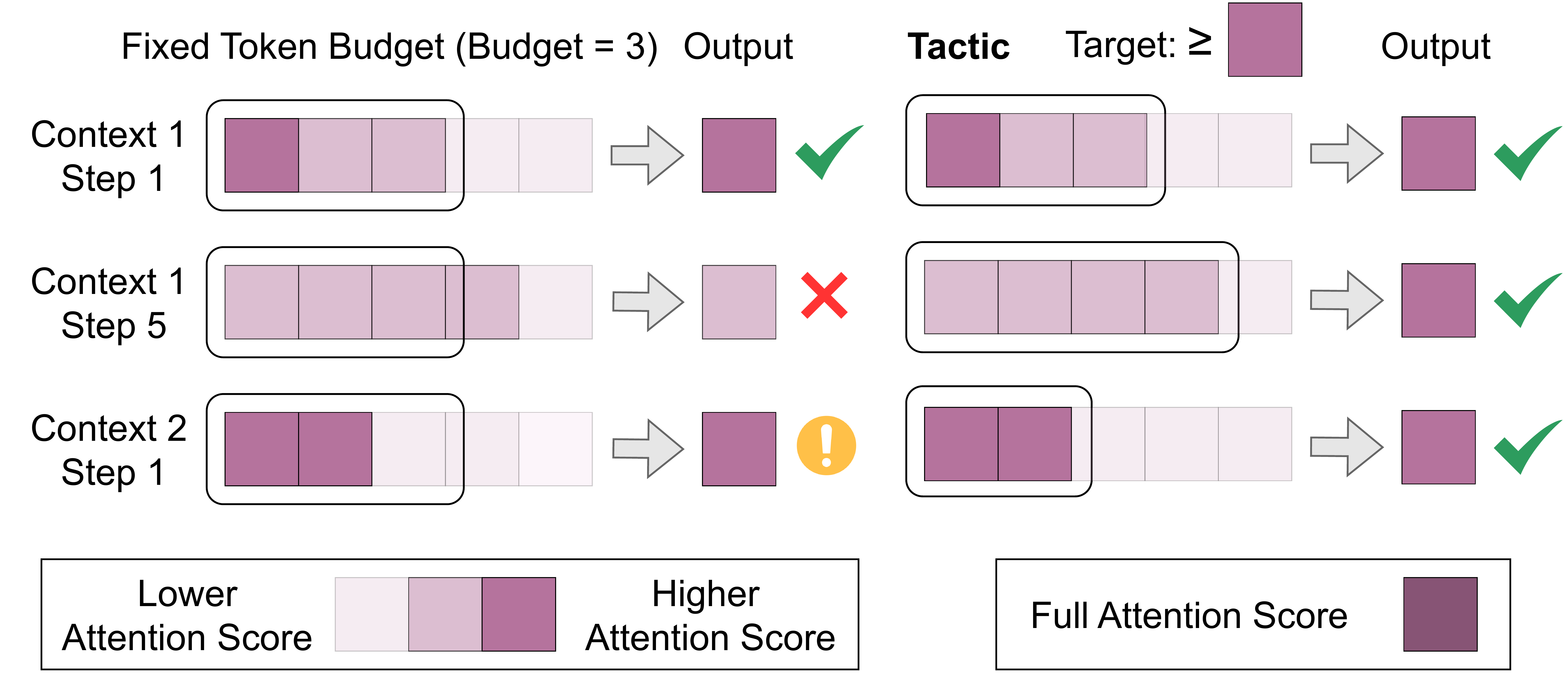} 
    \caption{Comparison between fixed-budget-based methods and \sys. Fixed-budget-based methods may select excessive tokens or have a large difference from full attention score. In contrast, \sys dynamically selects tokens to efficiently approximate full attention based on a cumulative attention score, considering variation of sparsity across different query tokens and contexts.}
    \label{fig:teaser}
\end{figure}

Alas, existing fixed budget-based methods have several shortcomings. Some methods employ a global fixed token budget~\cite{tang2024questqueryawaresparsityefficient,xiao2023streamingllm,zhang2023h2o}, not accounting for variations in attention sparsity across attention heads, and layers. In practice, some attention heads focus on significantly more tokens than others, and the level of sparsity fluctuates across layers. More adaptive methods~\cite{cai2024pyramidkvdynamickvcache,feng2024adakvoptimizingkvcache,ge2024modeltellsdiscardadaptive} attempt to distribute token budgets more effectively using calibration data or predefined rules, but they remain constrained by static allocation and cannot adapt to query tokens and contexts, often leading to suboptimal approximations in different cases.

To address the limitations of fixed-budget-based methods, we propose \sys, a sparsity-adaptive and
calibration-free post-training sparse attention mechanism that improves both the accuracy and efficiency of long-context LLM inference. \Cref{fig:teaser} shows a comparison between existing fixed budget-based methods and \sys. Instead of enforcing a fixed budget, \sys dynamically selects tokens starting from ones with the highest \as{} to ensure that their cumulative attention scores (where \as{} represents the softmax output of the Query-Key product) reach a target fraction of the full \as{}.

Dynamic and selective accumulation of \as{}s offers two key advantages. First, it provides inherent flexibility---\sys selects fewer tokens in high-sparsity cases and more in low-sparsity cases without requiring calibration. Second, since \as{}s are multiplied by \( V \) vectors with similar norms, and the selected \as{}s cumulatively reach at least a fraction of the total \as{}, a cumulative \as{} target guarantees, unlike token budgets in prior works, a bounded difference between sparse and full attention (see \Cref{subsec:analysis-attn-approx} and \Cref{sec:appendix-bound-proof}).

However, efficiently selecting tokens to reach a certain fraction $P$ of cumulative \as{} is challenging. To minimize the number of tokens selected (i.e., loads from memory), the optimal way is to select tokens following a descending order of \as{} until the cumulative \as{} surpasses $P$. Thus, similar to prior works, efficiently sorting tokens by their contribution to the cumulative \as{} is crucial for \sys. However, unlike fixed budget-based methods that simply stop at a fixed token count, \sys must track cumulative \as{} in real time, requiring the exact \as{} values for each token, making the selection process more complex.

To approximate optimal token selection, \sys introduces two key techniques: clustering and distribution fitting. First, to efficiently sort tokens, \sys clusters similar tokens to reduce computational overhead. However, we observe that positional proximity, which is used for clustering tokens by prior work~\cite{tang2024questqueryawaresparsityefficient}, does not necessarily guarantee similarity in Key vectors, which are fundamental to attention computation. Since attention operates on Query-Key interactions rather than token positions, \sys groups tokens using K-means clustering based on Key-vector similarity (i.e., vector distance). During decoding, \sys approximates the sorted list of tokens by sorting clusters based on the similarity between Query vectors and cluster centroids. After approximating token sorting,  \sys estimates the \as{} for each token by leveraging the observation that attention scores follow a smooth distribution. Using \df, \sys effectively keep track of attained cumulative \as{} to determine the end of selection.

By loading only the cluster centroids along with a small sampled subset of tokens ($\sim2.5\%$ of the KV cache size in practice), \sys efficiently selects the most critical tokens that reach the target cumulative \as{}. To balance efficiency and accuracy, \sys performs full attention on newly generated tokens and updates the clustering every fixed number of decoding steps (e.g., 2048).

Our experiments show that \sys achieves superior and consistent accuracy compared to existing algorithms including Quest~\cite{tang2024questqueryawaresparsityefficient}, PyramidKV~\cite{cai2024pyramidkvdynamickvcache} and Ada-KV~\cite{feng2024adakvoptimizingkvcache}, offering a more effective solution for long-context LLM inference in accuracy-sensitive applications. \sys achieves up to \attnspeedup decode attention speedup, which leads to \etoespeedup end-to-end speedup.

In summary, we contribute the following:
\begin{itemize}
    \item A detailed analysis of the dynamic nature of attention sparsity across heads, layers, queries, and contexts.
    \item \textbf{\sys}, a sparsity-adaptive attention algorithm that uses clustering and \df to dynamically determine the token budget for achieving cumulative \as{} targets. %
    \item A comprehensive evaluation of \sys{}, demonstrating \sys{} consistently achieves high accuracy and significant speedup.
\end{itemize}

\section{Background}
\subsection{Large Language Models}

LLMs consist of transformer blocks, each with an attention and a feed-forward module. In the attention module, input embeddings are projected into Query ($Q$), Key ($K$), and Value ($V$) vectors. For a given $Q$ vector, attention weights  $w_i$ for the i-th token are computed as $(\frac{QK^\top}{\sqrt{d}})_i$, then normalized via softmax to obtain \as{}s. These \as{}s are multiplies by the $V$ vectors to produce the output (\Cref{eq:attn}). This result is then processed by the feed-forward module, with residual connections and layer normalization refining the final token representation.

\begin{equation}
    \label{eq:attn}
    O=\text{softmax}\left(\exp\left(\frac{QK^\top}{\sqrt{d}}\right)\right)V
\end{equation}
At the request level, LLM inference consists of two phases: the prefill phase and the decode phase. In the prefill phase, all input tokens are processed simultaneously to generate $Q$, $K$, and $V$ vectors, with the $K$ and $V$ vectors stored in the KV-cache to avoid recomputation. In the decode phase, only the last generated token is processed, with its  $Q$, $K$, and $V$ vectors computed. The current $Q$ vector interacts with the cached $K$ and $V$ vectors to generate the output.

Unlike prefill, the decode phase is executed for each generated token, making it the primary bottleneck in inference for many workloads. For instance, in summarization tasks with 64K-token input documents, generating just 1024 tokens can take up to four times longer than the prefill phase.  

\subsection{Long-Context Models}

The recent increasing demand for long-context models has driven advancements in extending the context window of LLMs. Techniques like Rotary Position Embeddings~\cite{su2023roformerenhancedtransformerrotary} have enabled a significant expansion in context length, such as increasing LLaMA-2's context window to 32K in LongChat~\cite{longchat2023} and 128K in Yarn-Llama-2~\cite{peng2023yarn}, and even 1 million recently~\cite{liu2024scalinglawsropebasedextrapolation}. However, as context length grows dramatically, loading the KV cache becomes an increasingly significant bottleneck during token generation, which can account for over half of the total decode time~\cite{tang2024questqueryawaresparsityefficient}.

\section{Analysis}
\begin{figure}[t]
    \centering
    \includegraphics[draft=false, width=0.8\columnwidth]{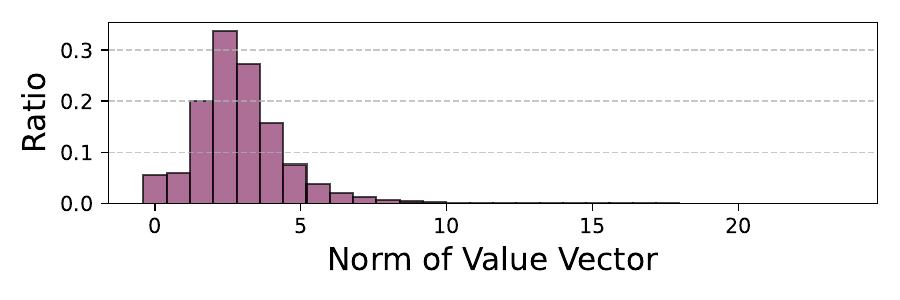} 
    \caption{The distribution of  $||V||$ across different layers, heads, and decoding tokens. The results indicate that $||V||$ values are concentrated within a very narrow range.}
    \label{fig:distribution_v}
\end{figure}

\begin{figure}
    \centering
    \begin{subfigure}[b]{0.49\columnwidth}
        \centering
         \includegraphics[draft=false, width=\linewidth]{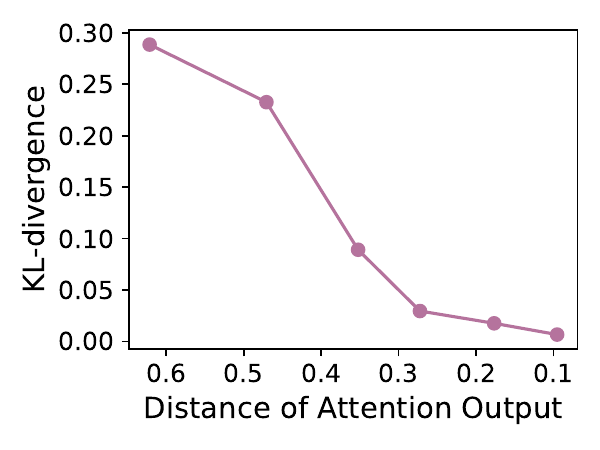}
        \caption{}
        \label{fig:kld_vs_attn_dis}
    \end{subfigure}
    \hfill
    \begin{subfigure}[b]{0.49\columnwidth}
        \centering
         \includegraphics[draft=false, width=\linewidth]{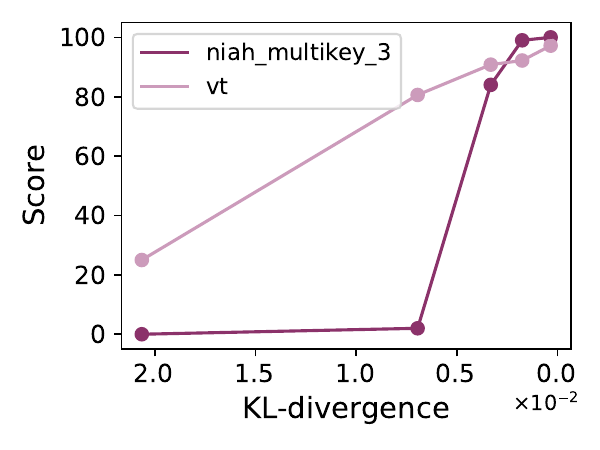}
        \caption{}
        \label{fig:score_vs_kld}
    \end{subfigure}
    \vspace{-0.1in}
    \caption{Comparison of KL-Divergence with attention distance (a) and its relation with downstream task scores (b).}
    \label{fig:combined}
\end{figure}

In this section, we first analyze the intrinsic sparsity in attention and how this sparsity affects downstream task performance (\Cref{subsec:analysis-intrinsic-sparsity} \& \Cref{subsec:analysis-attn-approx}).
We then present the drawbacks of existing fixed token budget approaches and propose cumulative \as{} as the new target (\Cref{subsec:analysis-cumulative-as} \& \Cref{subsec:analysis-challenge-as}).
Finally, we illustrate the challenges of applying cumulative \as{} and propose clustering and distribution-fitting techniques as the solution (\Cref{subsec:analysis-clustering} \& \Cref{subsec:analysis-curve-fitting}).

\subsection{Intrinsic Sparsity in Self-Attention Mechanisms}
\label{subsec:analysis-intrinsic-sparsity}

In the decode phase, for one request, assuming there are $n$ previous tokens, the attention formula in  \Cref{eq:attn} can be rewritten as

\begin{equation}
    o = \sum_{i=1}^{n}s_iv_i,~~s_i=\frac{\exp(\frac{qk_i^\top}{\sqrt{d}})}{\sum_{i=1}^{n}\exp(\frac{qk_i^\top}{\sqrt{d}})}.
\end{equation}

\fig{fig:distribution_v} shows that the distribution of $\|v_i\|$ for each token has a small variance. Thus, the contribution of the token is mostly determined by $s_i$. 
Due to the exponential term $\exp(\frac{qk_i^\top}{\sqrt{d}})$, only a small subset of tokens has a significant impact on the model output~\cite{zhang2023h2o,xiao2023streamingllm}, indicating that the attention operation in LLMs is inherently sparse, which motivates the possibility of only loading a subset of tokens to approximate the attention output and incur lower computational overhead.

\subsection{Rethinking Attention Approximation}
\label{subsec:analysis-attn-approx}
The sparse attention methods can be formulated as
\begin{equation}
    \tilde{o}(I) =  \sum_{i\in I}\tilde{s_i}v_i,~~\tilde{s_i}=\frac{\exp(\frac{qk_i^\top}{\sqrt{d}})}{\sum_{i\in I}\exp(\frac{qk_i^\top}{\sqrt{d}})}
\end{equation}
where $I$ is the index set of tokens selected by the sparse attention method, $I\subset[n]$. The distance between $o$ and $\tilde{o}(I)$ can be formulated as 
\begin{equation}
    \epsilon(I)=\|o-\tilde{o}(I)\|.
\end{equation}

\begin{figure*}
    \centering
    \begin{subfigure}[b]{0.32\linewidth}
        \centering
         \includegraphics[draft=false, width=\linewidth]{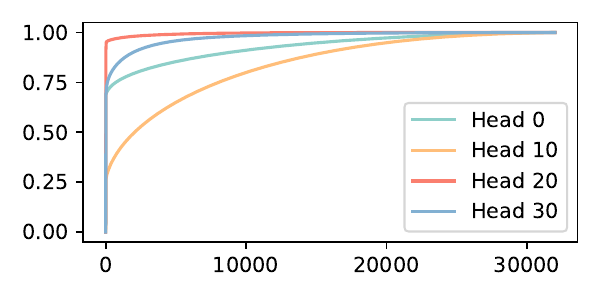}
        \caption{}
        \label{fig:headvar}
    \end{subfigure}
    \hfill
    \begin{subfigure}[b]{0.32\linewidth}
        \centering
         \includegraphics[draft=false, width=\linewidth]{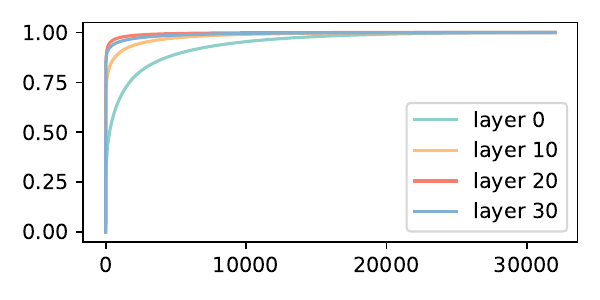}
        \caption{}
        \label{fig:layervar}
    \end{subfigure}
    \begin{subfigure}[b]{0.32\linewidth}
        \centering
         \includegraphics[draft=false, width=\linewidth]{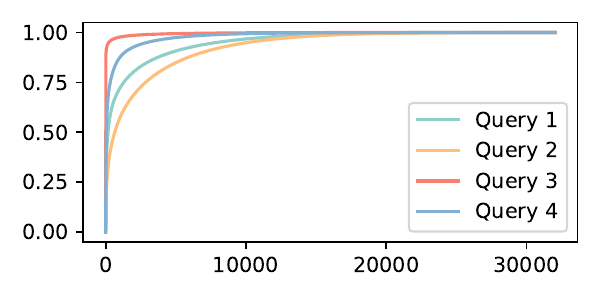}
        \caption{}
        \label{fig:queryvar}
    \end{subfigure}
    \vspace{-0.1in}
    \caption{Variation in sparsity across attention heads (a), model layers (b), and query tokens (c).}
    \label{fig:combined}
\end{figure*}

\begin{figure}[t]
    \centering
    \includegraphics[draft=false, width=\columnwidth]{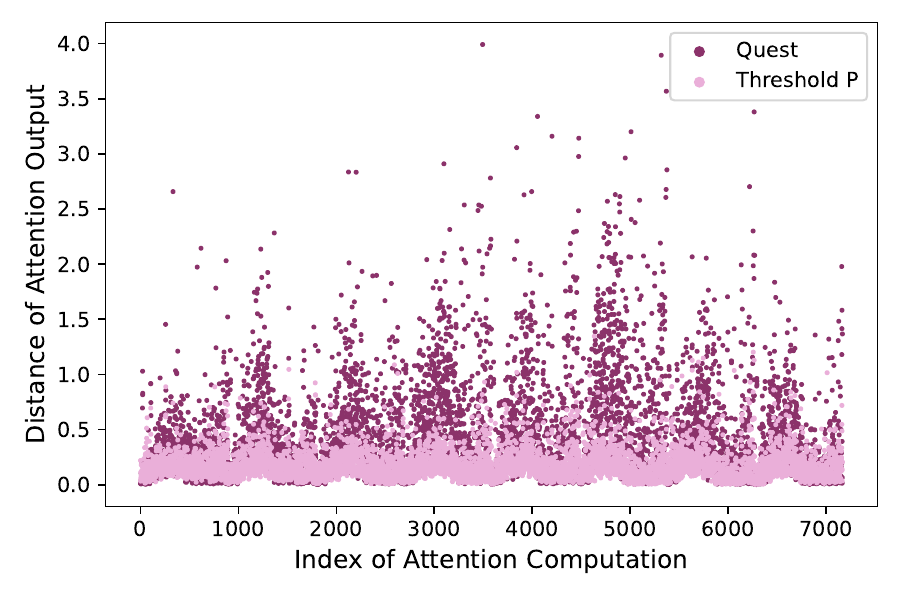}

    \caption{Distance of attention output to full attention of \quest~\cite{tang2024questqueryawaresparsityefficient} and setting cumulative \as{} threshold P, measured with Llama3.1-8B-Instruct model. Each dot represents the distance of one attention computation identified by (head index, layer index, decode step). }
    \label{fig:attn_dis_vary}
\end{figure}

Intuitively, a smaller distance between $o$ and $\tilde{o}(I)$ reduces the difference between the output token distributions of sparse attention and full attention. We demonstrate this by measuring the Kullback–Leibler (KL) divergence, a statistical metric that quantifies the difference between probability distributions, on the output logits as a function of attention distance. \Cref{fig:kld_vs_attn_dis} shows that decreasing attention distance consistently lowers KL-divergence, meaning the sparse attention output more closely resembles the full attention output. This alignment in token distributions also improves downstream task accuracy, as demonstrated in \Cref{fig:score_vs_kld} for tasks in the RULER benchmark, which is widely used to assess the long-context abilities of a model.

Therefore, the goal of sparse attention is to find an index set $I$ that minimizes $\epsilon(I)$ under some constraint of $|I|$.

\subsection{Fixed Token Budget Approaches Lead to Accuracy Variations}
\label{subsec:analysis-fixed-token-budget}
Several methods have been proposed to choose a small set of tokens $I$ minimizing the distance $\epsilon(I)$ between full and approximate attention. Some of the work, including Quest~\cite{tang2024questqueryawaresparsityefficient}, uniformly chooses tokens across attention heads and layers. These results in a large variance of $\epsilon(I)$, as shown in \fig{fig:attn_dis_vary}. This variance stems from the intrinsic sparsity difference across heads and layers. As illustrated in \fig{fig:headvar}, attention heads exhibit distinct sparsity patterns. Some heads display a more uniform distribution of $s_i$ (retrieval heads), whereas others are dominated by a few high-magnitude $s_i$ values (streaming heads). When a fixed number of tokens $|I|$ is selected per head, it leads to inefficiencies—allocating excessive tokens to streaming heads while introducing significant estimation errors in retrieval heads. Similarly, \fig{fig:layervar} highlights variation in sparsity across layers, where earlier layers exhibit lower sparsity compared to later ones, similarly making it inefficient to select a fixed number of tokens from different layers.

Motivated by the diversity of sparsity patterns across heads and layers, some works, including AdaKV\cite{feng2024adakvoptimizingkvcache} and PyramidKV\cite{cai2024pyramidkvdynamickvcache}, fix the total budget $|I|$ but use calibration data or assumptions to statically assign different budgets to different layers and heads. However, as we show in \fig{fig:queryvar}, the sparsity of particular heads varies significantly depending on the query token. For example, in the model output \textit{``The Answer is ...''}, the token \textit{``Answer"} attends to far fewer tokens compared to \textit{``is"}. This is because \textit{``Answer"} relies primarily on local information to generate \textit{``is"}, whereas \textit{``is"} requires broader context to produce the subsequent answer. Thus, relying on static partitioning of a fixed token budget also falls short of maintaining a consistent low attention distance $\epsilon(I)$.

\subsection{Cumulative \AS{}: A More Robust Target for Sparse Attention}
\label{subsec:analysis-cumulative-as}

The key drawback of existing work is the reliance on a fixed total token budget, making it hard to adapt to sparsity variations. Instead, we propose directly using the cumulative \as{} of tokens in $I$ to guide token selection.

Specifically, we define $p(I)$ as the cumulative \as{} of tokens in $I$, which is
\begin{equation}
    p(I)=\sum_{i\in I}s_i=\frac{\sum_{i\in I}\exp(\frac{qk_i^\top}{\sqrt{d}})}{\sum_{i=1}^{n}\exp(\frac{qk_i^\top}{\sqrt{d}})}
\end{equation}

These cumulative \as{} targets offer two key advantages over fixed token budgets. First, they inherently adapt to sparsity variations without requiring assumptions or calibration data. Less sparse heads, layers, query tokens, and contexts naturally require more tokens to reach a given cumulative \as{} than sparser ones. Second, targeting cumulative \as{} provides a theoretical guarantee on attention distance. Specifically, the attention distance is bounded by  

\begin{equation}
    \epsilon(I)\leq 2(1-p(I))\max_i\|v_i\|.
\end{equation}  

A detailed proof is provided in \Cref{sec:appendix-bound-proof}. Since value vectors $V$ have similar norms across tokens (\fig{fig:distribution_v}) , setting a threshold $P$ (typically close to 1.0) for $p(I)$ establishes a tight upper bound on $\epsilon(I)$. Identifying the minimal index set $I$ that satisfies $p(I) \geq P$ reduces the variance of the attention approximation error, as shown in \fig{fig:attn_dis_vary}. This improved attention distance approximation directly enhances downstream task performance, as demonstrated in \Cref{subsec:analysis-attn-approx} .

\subsection{Challenges of Attaining Cumulative \AS{}s}
\label{subsec:analysis-challenge-as}

Identifying the minimal subset of tokens that achieve a target cumulative \as{} is a challenging task. The optimal way is to select tokens following a descending order of \as{} until the cumulative \as{} surpasses the target value. Therefore, like prior approaches, \sys must rank tokens by \as{} to minimize the number of tokens needed to reach the desired cumulative \as{}. However, unlike previous methods, \sys also requires the \as{} values for each token to track the cumulative sum of selected tokens in real-time. This process involves two key components: (1) computing the sum of attention intermediate values, $\exp(qk^\top/\sqrt{d})$, for the selected token set $I$, and (2) computing the total sum of $\exp(qk^\top/\sqrt{d})$ used for normalization. Additionally, this estimation must be computationally efficient, as it lies on the critical path during decoding.

\subsection{Sorting Tokens via Clustering}
\label{subsec:analysis-clustering}
\begin{figure}[t]
    \centering
    \includegraphics[draft=false, width=\columnwidth]{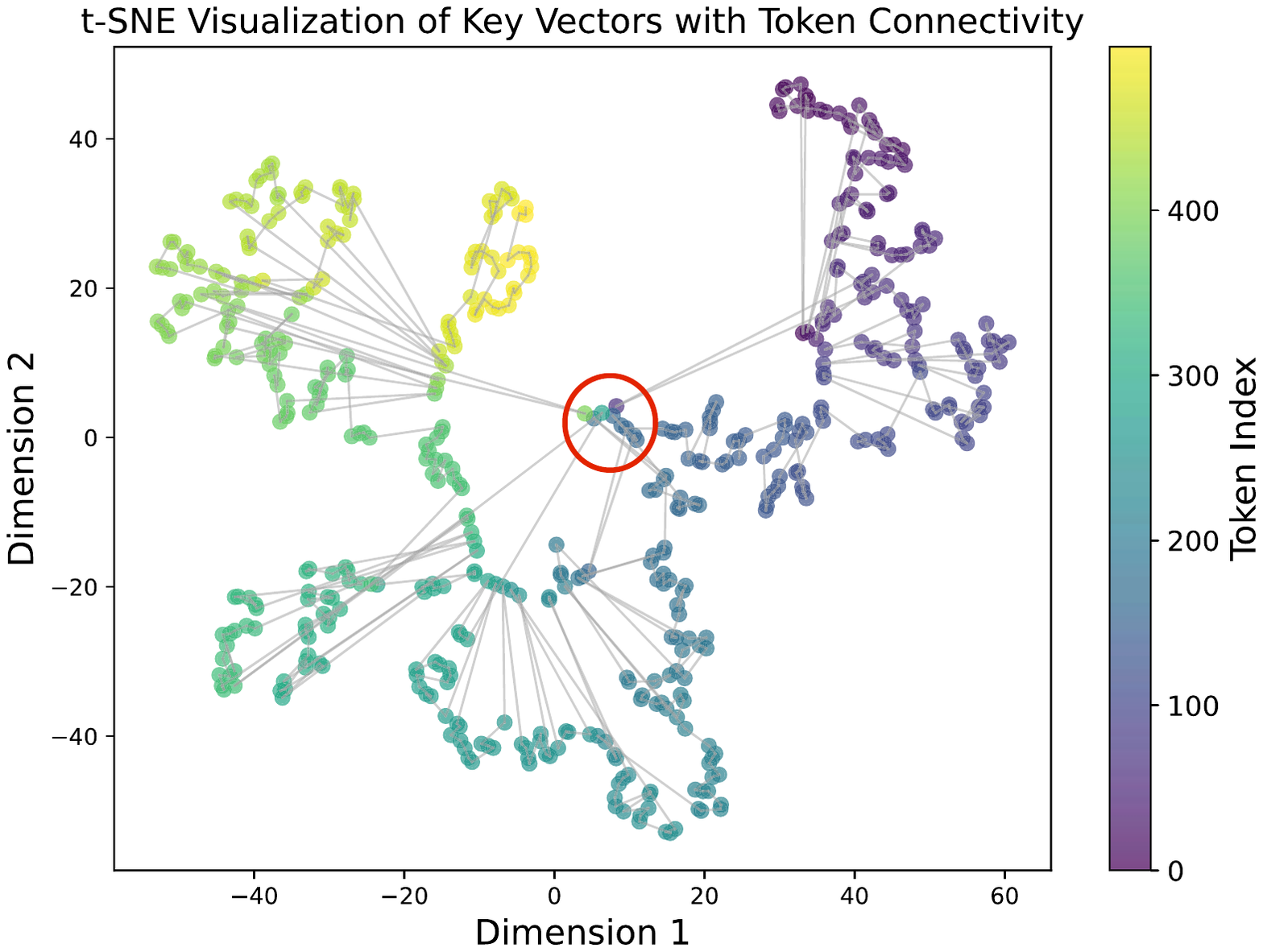} 
    \caption{t-SNE visualization of 500 Key vectors. Consecutive tokens are connected by lines. There are significant jumps and discontinuities even if tokens are consecutive. This indicates that adjacent tokens may not have similar K-vectors. Nonetheless, at the center of the figure (circled), K-vectors from different segments of the text show high similarity.}
    \label{fig:scatter_key}

\end{figure}

Similar to prior works, \sys groups tokens to reduce computational overhead. However, existing methods rely on positional order, assuming consecutive tokens share similar attention patterns~\cite{tang2024questqueryawaresparsityefficient}. As shown in \Cref{fig:scatter_key}, this is suboptimal since Key vectors of consecutive tokens are often scattered in the embedding space, meaning positional proximity does not imply similarity in attention behavior. Moreover, modern attention kernels efficiently handle non-contiguous KV-cache access, making positional grouping unnecessary. Instead, \sys applies K-means clustering to group tokens based on Key-vector similarity, then ranks them using the dot product between Query vectors and cluster centroids, ensuring selection aligns with actual attention behavior.

The runtime performance overhead of cluster-based sorting is $\frac{1}{2 \times \text{Average Cluster Size}}$,   compared to full attention\footnote{The term 2 comes from clustering being only performed on K-cache, while the KV cache is twice as large as K-cache.}, which in practice is below $2\%$.

We validate the results of clustering by showing the ground truth \as{} of tokens after sorting them based on clustering and estimation in \fig{fig:distribution_curve}. Despite rare spikes, clustering-based sorting gives a high-fidelity approximation of full attention-based token ordering.
\subsection{Estimating \AS{} via Distribution Fitting}
\label{subsec:analysis-curve-fitting}
\begin{figure}[t]
    \centering
    \includegraphics[draft=false, width=\columnwidth]{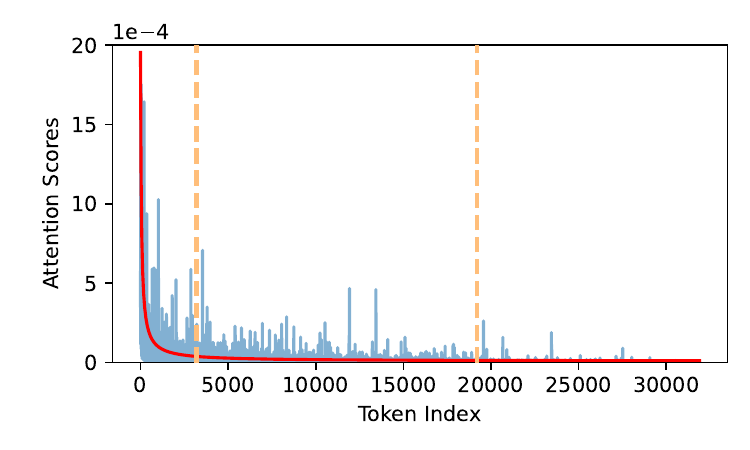} 
    \caption{The distribution of attention scores after cluster-based sorting for one request in PG19 dataset using Llama3.1-8B-Instruct model. Despite some variations, the overall trend closely aligns with the function y = $\frac{a}{x}+b$.}
    \label{fig:distribution_curve}
\end{figure}
\vspace{-0.1in}

While clustering effectively sorts tokens by \as{}, it introduces large errors when estimating absolute \as{} values. This occurs because the cluster centroid represents the center of tokens, but due to non-linearity, its \as{} does not accurately reflect the average \as{} of individual tokens. Thus, \sys requires a more precise approach to estimating \as{}. We observe that after partial sorting, the \as{} distribution follows a consistent pattern across heads, layers, and contexts. For example, as shown in \fig{fig:distribution_curve}, the \as{} is high for a few tokens and then smoothly decreases, forming a long-tail distribution. This structure suggests that function fitting can be used to estimate \as{}. Despite outliers at the beginning of the curve, sampling tokens along the distribution allows accurate parameter estimation, enabling precise \as{} predictions.

\section{Methodology}
\label{sec:method}
\begin{figure}[t]
    \centering
    \includegraphics[draft=false, width=0.9\columnwidth]{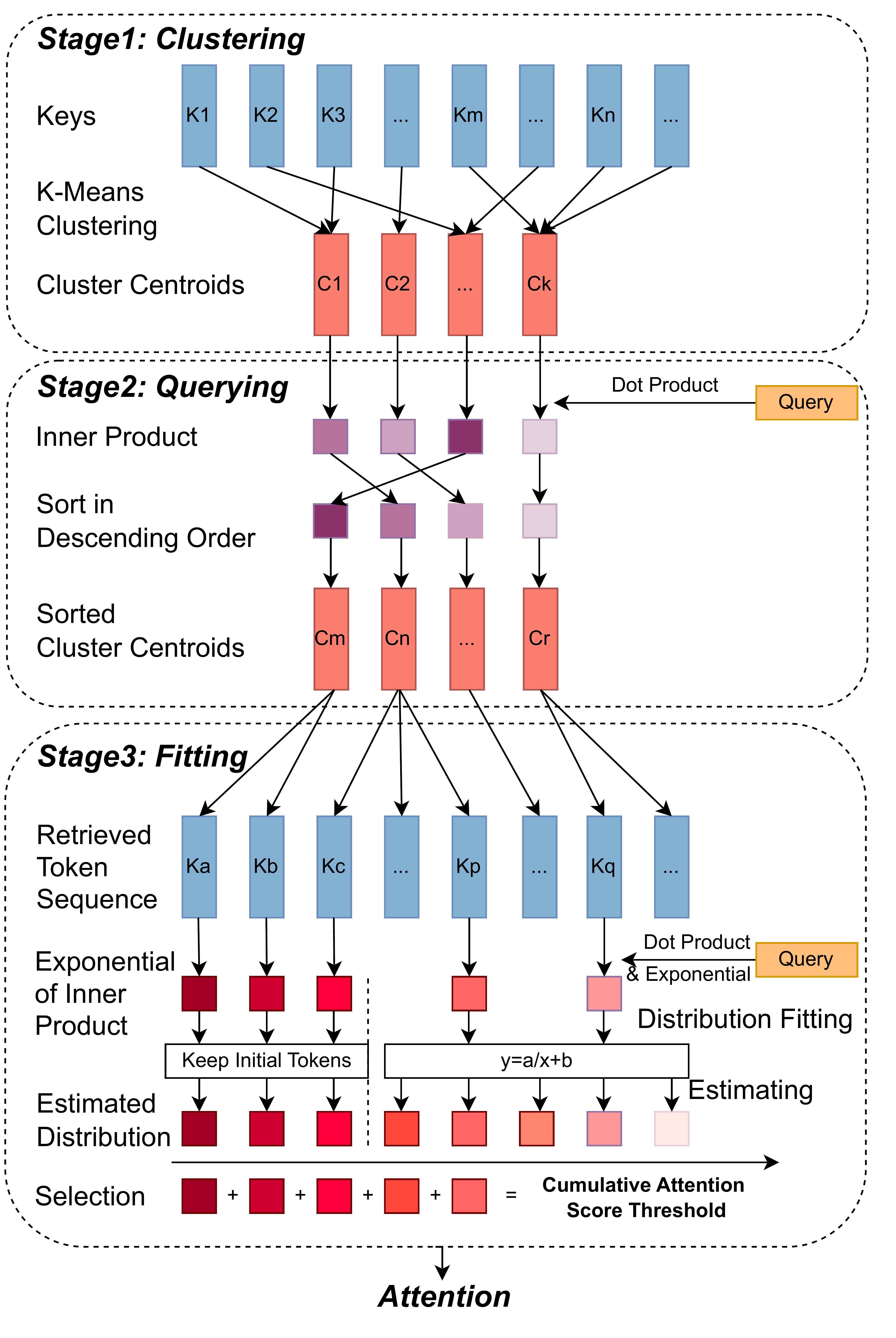}
    \caption{ The overall workflow of \sys. \sys operates in three stages to achieve low overhead adaptive sparse attention.
    }
    \label{fig:Methodology}
    \vspace{-0.1in}
\end{figure}

\subsection{Algorithm Overview}
\label{sec:methodology:overview}
\Cref{fig:Methodology} provides an overview of \sys's workflow. During prefill,  \sys performs K-means clustering on key vectors to group similar tokens. During decode, \sys ranks tokens based on the dot product between cluster centroids and the current query vector. \sys then models the distribution of \as{} with a fitted curve and determines the tokens to meet the desired cumulative \as{} threshold. After token selection, \sys handles the Group Query Attention (GQA) and then performs the attention using FlashInfer~\cite{ye2025flashinfer}.

\subsection{Clustering}
To organize tokens for efficient sorting, \sys performs K-means clustering on the key vectors for each head in every layer during the prefill phase. We empirically choose the average cluster size to be $32$ to balance accuracy and efficiency. Clustering begins by randomly sampling $\text{\textit{SeqLen}} / \text{\textit{Average cluster size}}$ data points as the initial cluster centroids.\footnote{Note that neither multiple initializations nor K-Means ++ initialization drastically improves the clustering quality, and in fact leads to high-performance overhead.}In each iteration, the distance between K-vectors and centroids is computed and the token will be assigned to the nearest cluster. After the assignment step, the centroids are updated as the mean of the key vectors assigned to each cluster. This process repeats until convergence or until a maximum of 10 iterations is reached\footnote{More iterations do not improve the quality of clustering.}. 

\subsection{Querying}

Once the tokens are organized into clusters, \sys identifies critical clusters for a given query vector $Q$ in the decode phase. The criticality of each cluster is determined by the dot product between $Q$ and each cluster centroid\footnote{Compared to distance, dot product directly relates to the attention score, which is more accurate.}. This process produces a sequence of clusters sorted by the criticality, from which we can derive a partially sorted token list. 

\subsection{Fitting Attention Score Distribution}

The next step of \sys is to determine the token budget required to meet the cumulative \as{}. \sys models the distribution of the exponential values of the dot products ($\exp(\frac{QK^\top}{\sqrt{d}})$) for each token using a lightweight function \( y = \frac{a}{x} + b \), where \( a \) and \( b \) are parameters to be determined and \( x \) is the position in the sorted list of tokens.
To estimate these parameters, we select two segments of the tokens in the middle of the curve (e.g., 10\% and 60\% of all the tokens), and calculate the average of tokens within each segment (as labeled in \fig{fig:distribution_curve}). Using these two data points, we can solve for \( a \) and \( b \), which provides an estimation of \as{} for all tokens.

However, initial tokens are often outliers and cannot be accurately described by the curve. Moreover, these tokens feature high \as{}, and thus a bad estimation would cause high deviations of estimated cumulative \as{} which affects the accuracy of \sys.
Luckily, we observed that this only happens within 1-2\% of total tokens. Therefore, \sys directly calculates the exponential values of the dot products for these tokens. A detailed description of the Distribution Fitting stage is provided in \Cref{alg:token_selection}.

\subsection{Taking Union for Group Query Attention models}

Modern models use Grouped Query Attention (GQA) to reduce the KV cache size \cite{llama_3}, where multiple query heads share a single KV head. However, loading KV heads separately for each query head is inefficient. To optimize this, query heads within the same group are batched. A challenge arises when using sparse attention, as different query heads may select to attend to different KV tokens. Finding the minimal set of KV tokens that satisfies the cumulative attention scores (\as{}) across all query heads is NP-hard. To address this, \sys simplifies the problem by taking the union of selected tokens across all query heads and loading them at once, ensuring that each head retains the KV tokens it requires to perform attention while reducing repetitive loading.

\subsection{Attention on Selected Tokens}
Finally, \sys performs actual attention for selected tokens using FlashInfer~\cite{ye2025flashinfer}.
Notably, variations in sparsity across different heads cause an imbalanced attention workload. Traditional implementations primarily address imbalances across varying request lengths but struggle to handle head-level imbalance efficiently. To address this, \sys divides each request into subrequests. Each subrequest processes a KV head and its corresponding Query head, with sequence length determined by the tokens selected for each KV head. This transforms head-level imbalance back into sequence-level imbalance, which Flashinfer handles efficiently.

\section{Experiments}

\subsection{Setting}

We evaluate \sys for both accuracy and efficiency. We use two models: Llama-3.1-8B-Instruct~\cite{grattafiori2024llama3herdmodels}, a widely used model with Grouped-Query Attention; and MegaBeam-Mistral-7B-512k~\cite{megabeam-mistral-7B-512k-2024}, an extended version of Mistral-7B-Instruct-v0.2 with a 512k token context window.

For accuracy evaluations, we use the PG19 language modeling dataset \cite{rae2019compressivetransformerslongrangesequence}, six tasks from the LongBench dataset\cite{bai-etal-2024-longbench}, including HotpotQA\cite{yang-etal-2018-hotpotqa}, TriviaQA\cite{joshi-etal-2017-triviaqa}, MultifieldQA\cite{bai-etal-2024-longbench},  NarrativeQA\cite{kocisky-etal-2018-narrativeqa}, Qasper\cite{dasigi-etal-2021-dataset}, and Musique\cite{bai-etal-2024-longbench}. Additionally, we conduct experiments on the RULER benchmark\cite{hsieh2024rulerwhatsrealcontext}, using 50 examples for each dataset. We compare \sys with the most popular fixed token budget KV cache eviction algorithms, \quest~\cite{tang2024questqueryawaresparsityefficient}, \pyramid~\cite{cai2024pyramidkvdynamickvcache} and \adakv~\cite{feng2024adakvoptimizingkvcache}. To ensure consistency, we set the page size in Quest and the cluster size in our method to 16. Both \adakv and \pyramid follow the configuration
settings outlined in \cite{feng2024adakvoptimizingkvcache}, including an observation window size of 32 and a max pooling kernel size of 7. 
For the clustering process, we limit the maximum number of iterations to 10.

For efficiency evaluations, we perform the evaluation on Nvidia Ada 6000 GPUs with CUDA 12.4 compared with full attention using Flashinfer~\cite{ye2025flashinfer}.

\subsection{Accuracy Evaluation}
\begin{figure}[t]
    \centering
    \includegraphics[draft=false, width=\columnwidth]{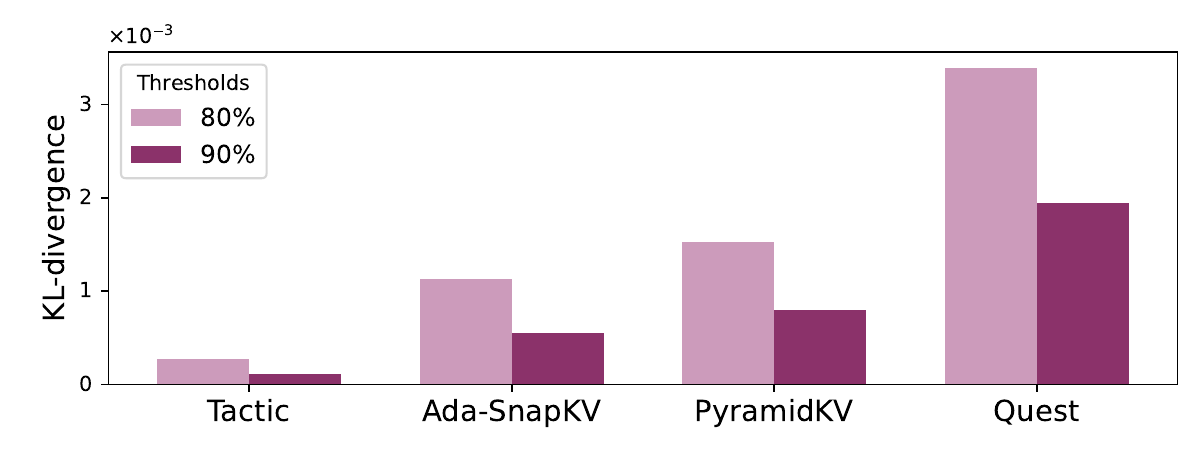} 
    \caption{KL-Divergence with full attention evaluation of \sys and other baseline methods on the PG19 dataset. \sys maintains the most accurate output in two configurations.}
    \label{fig:kld}
    \vspace{-0.1in}
\end{figure}

\begin{table}[t]
\centering
\resizebox{0.48\textwidth}{!}{
\begin{tabular}{l|c c c c c}
\toprule
\textbf{Threshold} & \textbf{Optimal} & \textbf{Cluster Optimal} & \textbf{\sys} & \textbf{Achieved} & \textbf{Success} \\
\hline
\midrule
\multicolumn{6}{c}{\textbf{\textit{Llama-3.1-8B-Instruct}}} \\
\hline
\hline
50\%  & 71 & 166 & 185 & 66\% & 92\% \\
60\%  & 122 & 271 & 294 & 72\% & 89\% \\
70\%  & 212 & 451 & 490 & 78\% & 86\% \\
80\%  & 394 & 802 & 890 & 84\% & 84\% \\
90\%  & 895 & 1723 & 1975 & 91\% & 86\% \\
\hline
\midrule
\multicolumn{6}{c}{\textbf{\textit{MegaBeam-Mistral-512k}}} \\
\hline
\midrule
50\%  & 71 & 166 & 185 & 66\% & 92\% \\
60\%  & 122 & 271 & 294 & 72\% & 89\% \\
70\%  & 212 & 451 & 490 & 78\% & 86\% \\
80\%  & 394 & 802 & 890 & 84\% & 84\% \\
90\%  & 895 & 1723 & 1975 & 91\% & 86\% \\
\bottomrule
\end{tabular}
}
\caption{Evaluation of number of tokens selected and ratio of cumulative \as{} achieved. The \textit{Optimal} method is to select tokens following a descending order o \as{}s. The \textit{Cluster Optimal} method is to select tokens following the order produced by clustering. Success means the ratio of cases achieve the threshold.}
\vspace{-5mm}
\label{tab:token_budget}
\end{table}

\begin{table*}[h!]
\centering

\small
\begin{tabular}{l c|c c c c c c}
\toprule
\hline
\textbf{Methods} & \textbf{Config} & \textbf{HotpotQA} & \textbf{TriviaQA} & \textbf{MultiFieldQA} & \textbf{Qasper} & \textbf{NarrativeQA} & \textbf{Musique} \\
\hline
\textit{Llama-3.1-8B-Instruct}  & Full  & 54.99  & 89.22  & 55.05  & 46.52  & 27.12  & 32.16 \\
Tactic  & 70\%  & 51.20  & \textbf{90.38}  & \textbf{52.98}  & \textbf{43.30}  & \textbf{30.39}  & \textbf{28.57} \\
PyramidKV  & 70\%  & \textbf{52.59}  & 89.57  & 43.26  & 27.50  & 21.44  & 25.34 \\
AdaKV   & 70\%  & 49.32  & 89.57  & 43.45  & 29.99  & 25.23  & 23.54 \\
Quest   & 70\%  & 45.43  & 77.42  & 49.96  & 38.09  & 24.77  & 24.78 \\
Tactic  & 90\%  & 53.57  & \textbf{90.61}  & \textbf{54.35}  & \textbf{44.20}  & 29.59  & \textbf{30.71} \\
PyramidKV  & 90\%  & 53.77  & 90.31  & 48.15  & 36.40  & 26.97  & 28.52 \\
AdaKV   & 90\%  & \textbf{54.05}  & 90.46  & 49.15  & 37.55  & 27.86  & 29.19 \\
Quest   & 90\%  & 49.37  & 80.38  & 52.42  & 42.41  & \textbf{30.21}  & 26.64 \\

\hline
\midrule
\textit{MegaBeam-Mistral-512k }   & Full  & 48.89  & 88.24  & 52.14  & 33.13  & 26.08  & 26.38 \\
Tactic  & 70\%  & \textbf{49.15}  & 87.89  & 50.50  & \textbf{32.37}  & \textbf{25.63}  & \textbf{25.85} \\
PyramidKV  & 70\%  & 42.21  & 85.77   & 36.74  & 21.23  & 19.31  & 19.93 \\
AdaKV   & 70\%  & 42.23  & 85.65  & 38.44  & 22.23  & 21.89  & 21.68 \\
Quest   & 70\%  & 48.90  & \textbf{88.13}  & \textbf{50.58}  & 30.78  & 23.88  & 24.65 \\
Tactic  & 90\%  & 49.59  & \textbf{89.16}  & 49.85  & \textbf{33.93}  & \textbf{26.31}  & \textbf{25.93} \\
PyramidKV  & 90\%  & 44.05  & 86.64   & 40.66  & 24.22  & 21.13  & 23.32 \\
AdaKV   & 90\%  & 44.80  & 86.80   & 42.80  & 22.51  & 22.46  & 24.86 \\
Quest   & 90\%  & \textbf{51.69}  & 88.49  & \textbf{51.81}  & 32.46  & 24.63  & 25.89 \\

\bottomrule

\end{tabular}
\caption{Evaluation Results on LongBench}
\label{tab:longbench}
\end{table*}

\subsubsection{Accuracy of Clustering \& Distribution Fitting}

To identify the minimal number of tokens to reach the threshold, \sys employs clustering and distribution fitting (explained in \Cref{sec:method}). We evaluate our method on the PG19 dataset, focusing on how well it aligns with the target cumulative \as{} and how many tokens it selects. We set specific \as{} thresholds and compare the actual cumulative score achieved by our method against two oracles: the global optimal, which sums tokens in the descending order of \as{}, and the clustering optimal, which sums \as{} from sorted clusters.
\Cref{tab:token_budget} presents the relative error between target and obtained cumulative \as{}, as well as the comparison of the number of tokens selected by these methods. \sys achieves the target threshold of cumulative \as{} on average with high success rates. Also, the values of \textit{Cluster Optimal} and \textit{\sys} are close, indicating that the distribution fitting presents an accurate estimation of number of tokens.

\subsubsection{Output Accuracy}
We assess the KL-divergence of model output probability distribution of \sys relative to the full attention under Top-K sampling using the PG19 test set\cite{rae2019compressivetransformerslongrangesequence}. 
We include all texts in PG19 with the number of tokens larger than 32k. In the prefill stage, we truncate the input to 32k tokens and feed it into the model. In the decode stage, we feed tokens one by one and collect the output logits of each decode step. We collect 32 decode steps in total. As shown in \Cref{fig:kld}, \sys achieves the most accurate output compared to all baselines.

\subsubsection{Accuracy for long-contexts tasks}
\begin{table*}[t]
\centering

\resizebox{0.7\textwidth}{!}{
\begin{tabular}{l|c c c c c c}
\hline
\textbf{P} & \textbf{HotpotQA} & \textbf{TriviaQA} & \textbf{MultiFieldQA} & \textbf{Qasper} & \textbf{NarrativeQA} & \textbf{Musique} \\
\hline
\multicolumn{7}{c}{\textbf{\textit{Llama-3.1-8B-Instruct}}} \\
\hline
70\%  & 959  & 761  & 774  & 629  & 1918 & 1229 \\
90\%  & 2298 & 1813 & 1559 & 1276 & 4254 & 2754 \\
\hline
\multicolumn{7}{c}{\textbf{\textit{MegaBeam-Mistral-7B-512k}}} \\
\hline
70\%  & 2126  & 1733  & 1399  & 1191  & 3017  & 2598  \\
90\%  & 3641  & 3048  & 2031  & 1546  & 5616  & 4384  \\
\hline
\end{tabular}
}
\caption{Average number of tokens selected by \sys for different cumulative \as{}s.}
\label{tab:token-count}
\end{table*}

\textbf{LongBench.}
We evaluate \sys on six LongBench tasks, namely, HotpotQA, TriviaQA, MultiFieldQA, Qasper, NarrativeQA, and Musique, spanning a wide range of scenarios such as single-document QA, multi-document QA, few-shot learning, and synthesis tasks. For each dataset, we first evaluate \sys by setting the cumulative \as{} threshold as 70\% and 90\%. The average number of tokens selected at each threshold serves as the token budget for evaluating \quest, \pyramid, and \adakv.

As shown in \tab{tab:longbench}, \sys consistently outperforms all other baselines. At a threshold of 90\%, \sys achieves performance close to full attention. We provide a detailed table of the average number of tokens selected by \sys across various thresholds, datasets and models in \tab{tab:token-count}, which is set as token budgets for baselines.

\begin{table}[t]
\centering

\caption{Performance comparison on RULER. Each score is computed by averaging accuracy of all tasks in RULER.}
\resizebox{\columnwidth}{!}{
\begin{tabular}{l c|c c c c|c}
\toprule
\hline
\textbf{Methods} & \textbf{Config} & \textbf{16K} & \textbf{32K} & \textbf{64K} & \textbf{96K} & \textbf{Avg.}\\
\hline
\textit{Llama-3.1-8B-Instruct} & Full & 91.3 & 86.0 & 85.2 & 85.0 & 86.8\\
\sys & 75\% & \textbf{90.9} & \textbf{85.5} & \textbf{83.4} & \textbf{78.9} & \textbf{84.7} \\
\pyramid & 75\% & 61.8 & 67.4 & 60.8 & 62.5 & 63.1 \\
\adakv & 75\% & 58.0 & 62.2 & 59.2 & 58.7 & 59.2\\
\quest & 75\% & 70.0 & 71.5 & 69.7 & 65.7 & 69.2 \\
\sys & 90\% & \textbf{90.3} & \textbf{84.9} & \textbf{82.8} & \textbf{80.5} & \textbf{84.6} \\
\pyramid & 90\% & 73.1 & 76.2 & 74.2 & 68.6 & 73.0\\
\adakv & 90\% & 72.7 & 76.4 & 74.3 & 68.7 & 73.0\\
\quest & 90\% & 85.8 & 81.9 & 79.8 & 70.5 & 79.5 \\
\hline
\midrule
\textit{Mega-Beam-Mistral-512k} & Full & 90.9 & 88.4 & 82.7 & 83.1 & 86.3\\
\sys & 75\% & \textbf{88.0} & \textbf{88.8} & \textbf{81.7} & \textbf{82.5} & \textbf{85.2} \\
\pyramid & 75\% & 80.2 & 79.0 & 75.2 & 74.0 & 77.1 \\
\adakv & 75\% & 80.6 & 78.1 & 75.4 & 73.6 & 76.8 \\
\quest & 75\% & 80.7 & 79.0 & 71.4 & 70.5 & 75.4 \\
\sys & 90\% & \textbf{90.3} & \textbf{88.0} & \textbf{81.0} & 82.6 & \textbf{85.4} \\
\pyramid & 90\% & 84.3 & 82.7 & 76.2 & 86.2 & 82.4 \\
\adakv & 90\% & 84.7 & 81.8 & 76.2 & \textbf{87.1} & 82.4 \\
\quest & 90\% & 81.1 & 81.3 & 73.5 & 79.7 & 78.9\\
\bottomrule
\end{tabular}
}
\label{tab:ruler}
\vspace{-0.1in}
\end{table}

\begin{table}[t]
\centering
\caption{Average number of tokens selected by \sys for different tasks, context lengths, cumulative \as{}s and models.}
\resizebox{\columnwidth}{!}{
\begin{tabular}{l|c|c|c|c|c|c|c|c}

\hline
\multicolumn{9}{c}{\textbf{\textit{Llama-3.1-8B-Instruct}}} \\
\hline
Task & \multicolumn{2}{c|}{16K} & \multicolumn{2}{c|}{32K} & \multicolumn{2}{c|}{64K} & \multicolumn{2}{c}{96K} \\
\cline{2-9}
 &  75\% & 90\%  & 75\%  & 90\% &  75\% & 90\% & 75\%  & 90\%  \\
\hline
NIAH\_Single 1 & 166 & 813 & 363 & 1567  & 456 & 2404  & 1790 & 3319 \\
NIAH\_Single 2 & 271 & 1289 & 534 & 2196  & 711 & 3209  & 2030 &  3940 \\
NIAH\_Single 3 & 171 & 1015  & 369 & 1820  & 507 & 3031  & 1068 & 3952 \\
NIAH\_Multikey 1 & 224 & 1052 & 449 & 1832  & 654 & 2792 & 978& 4438 \\
NIAH\_Multikey 2 & 399 & 1612 & 706 & 2533  & 981 & 3902  & 1405 & 5527  \\
NIAH\_Multikey 3 & 340 & 1428 & 567 & 2404  & 798 & 3990 & 1068 & 5062 \\
NIAH\_Multivalue & 246 & 1769 & 478 & 2679 & 692 & 4458  & 1025 & 8147 \\
NIAH\_Multiquery & 253 & 1524 & 465 & 2648  & 681 & 4830 & 915 & 6011 \\
FWE & 282 & 1572 &515 & 2693 & 778 & 4376  & 963 & 6202  \\
CWE & 443 & 1939 & 570 & 3036 & 655 & 4707 & 780 & 6731 \\
QA 1 & 329 & 1250  & 704 & 2565 & 852 & 3830  & 3417 & 5055 \\
QA 2 & 547 & 1484 & 1092 & 2263 & 1662 & 3455 & 2120 & 4276 \\
VT & 118 & 731 & 253 & 1413 & 269 & 2028 & 404 & 3068 \\
\hline
\hline
\multicolumn{9}{c}{\textbf{\textit{MegaBeam-Mistral-7B-512K}}} \\
\hline
Task & \multicolumn{2}{c|}{16K} & \multicolumn{2}{c|}{32K} & \multicolumn{2}{c|}{64K} & \multicolumn{2}{c}{96K} \\
\cline{2-9}
 &  75\% & 90\%  & 75\%  & 90\% &  75\% & 90\% & 75\%  & 90\%  \\
\hline
NIAH\_Single 1 & 1410 & 1176 & 2521 & 2354 & 4715 & 4512 & 7042 & 7271 \\
NIAH\_Single 2 & 1418 & 1665 & 2704 & 3364 & 5113 & 6732 & 7668 & 10472 \\
NIAH\_Single 3 & 3005 & 2032 & 5312 & 3800 & 10235 & 7327 & 15492 & 11090 \\
NIAH\_Multikey 1 & 1486 & 1661 & 2872 & 3415 & 5332 & 5983 & 7781 & 8767 \\
NIAH\_Multikey 2 & 1480 & 1639 & 2826 & 3031 & 5288 & 5956 & 8240 & 9279 \\
NIAH\_Multikey 3 & 2840 & 2039 & 5990 & 4306 & 11438 & 7741 & 17088 & 11292 \\
NIAH\_Multivalue & 2880 & 2225 & 5070 & 4006 & 10180 & 7463 & 16138 & 11853 \\
NIAH\_Multiquery & 3088 & 2165 & 5234 & 4075 & 9931 & 7551 & 14420 & 11164 \\
FWE & 2706 & 1809 & 5537 & 3685 & 10837 & 7157 & 16558 & 10818 \\
CWE & 4240 & 2526 & 7404 & 4869 & 13189 & 8802 & 18329 & 13215 \\
QA 1 & 1003 & 1502 & 2772 & 3132 & 5828 & 5685 & 6007 & 9150 \\
QA 2 & 724 & 1718 & 2124 & 3017 & 4853 & 6520 & 8598 & 8598 \\
VT & 2227 & 1409 & 4132 & 2619 & 8667 & 4719 & 14318 & 8671 \\
\hline

\end{tabular}
}

\label{tab:ruler-token-count}
\end{table}

\textbf{RULER.}
We evaluate \sys and baselines on all tasks in RULER~\cite{hsieh2024rulerwhatsrealcontext} with context length ranging from 16K to 96K.
As shown in the \tab{tab:ruler}, \sys consistently outperforms all baselines in each configuration in terms of average accuracy. Furthermore, at higher thresholds, \sys achieves similar accuracy to full attention, significantly higher than other methods. Similar as \tab{tab:token-count}, we provide a detailed table of token budgets used by baselines in \tab{tab:ruler-token-count}.

\subsection{Efficiency Evaluation}

We begin by analyzing the latency breakdown of \sys, focusing on token clustering during the prefill phase and attention computation for critical tokens during decoding (\Cref{sec:eval:breakdown}). Next, we evaluate \sys's end-to-end performance and its speed-up relative to full attention (\Cref{sec:eval:e2e}).

\subsubsection{Latency Breakdown}
\label{sec:eval:breakdown}

\textbf{Latency of clustering during prefill.}
We measure the time taken clustering for different sequence lengths in \Cref{tab:cluster_time}. We divide the clustering time into two steps. The first step is called distance calculation, where each K-vector computes its distance from the cluster centroids and assigns itself to the nearest cluster. The second step is called cluster update, where the centroids are updated based on the distances of K-vectors in the cluster.

We observe that, as the sequence length increases, the clustering time increases quadratically and is dominated by the distance calculation. However, large sequences also significantly increase the prefill time. Overall, across different sequence lengths, the clustering time always stays below 6\% of the prefill time.

\begin{table}[]
\centering
\caption{Comparison of clustering time and prefill time. Getting distance between K-vectors and cluster centroid dominates the clustering time. For all sequence lengths, clustering overhead is negligible compared to prefill. }
\vspace{0.1in}
\footnotesize
\resizebox{0.8\columnwidth}{!}{
\begin{tabular}{rrrr}
\hline
\textbf{SeqLen}  & \textbf{Get Distance (s)} & \textbf{Update (s)} & \textbf{Prefill (s)} \\ \hline
32768   & 0.15            & 0.01       & 5.66        \\
65536   & 0.61            & 0.01       & 15.80       \\
131072  & 2.72            & 0.02       & 49.53       \\

\end{tabular}
}
\label{tab:cluster_time}
\vspace{-0.1in}
\end{table}

\begin{figure}[t]
    \centering
    \includegraphics[draft=false,width=\columnwidth]{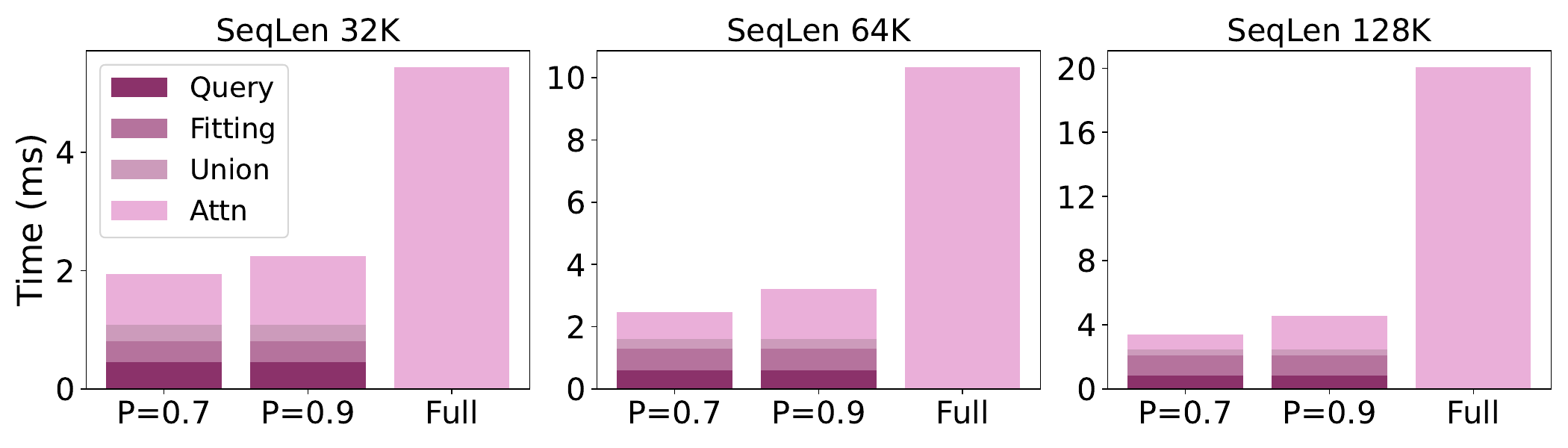}
    \caption{Latency breakdown of \sys in the attention of decode stage for different sequence lengths and different thresholds.}
    \label{fig:decoding_time}
\end{figure}
\begin{figure}[t]
    \centering
    \includegraphics[draft=false, width=\columnwidth]{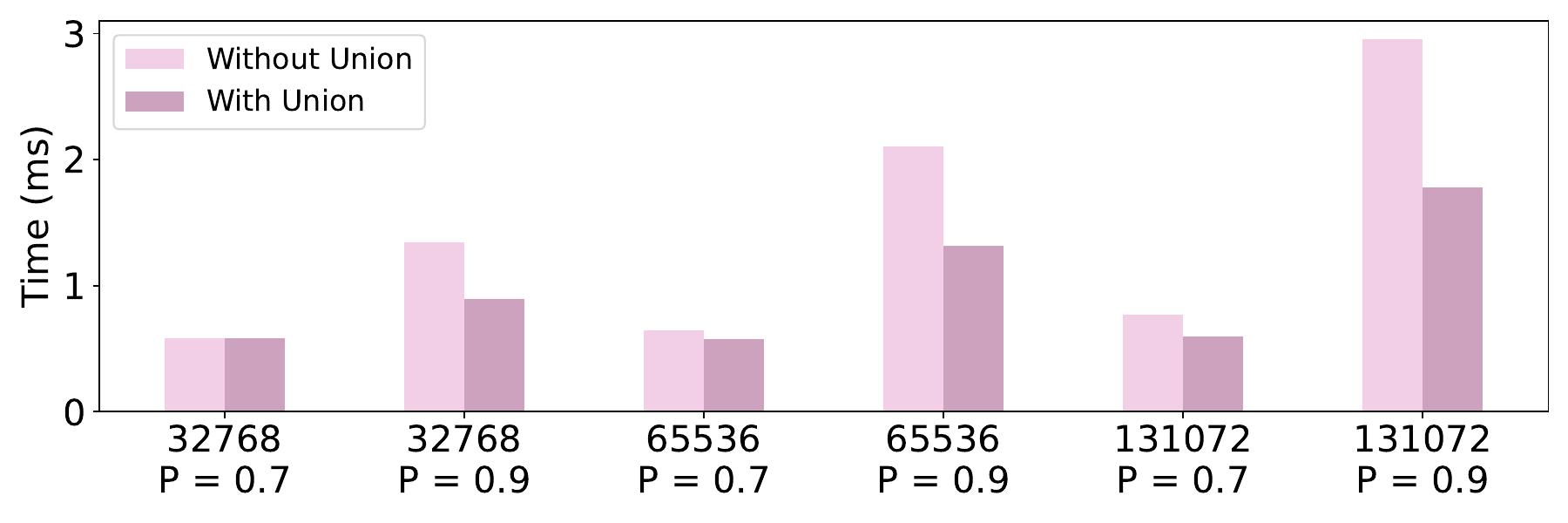}
    \caption{Ablation study on taking union for the GQA model. Taking union significantly reduce the attention time. }
    \label{fig:union}
\end{figure}
\begin{figure}[t]
    \centering
    \includegraphics[draft=false,width=\columnwidth]{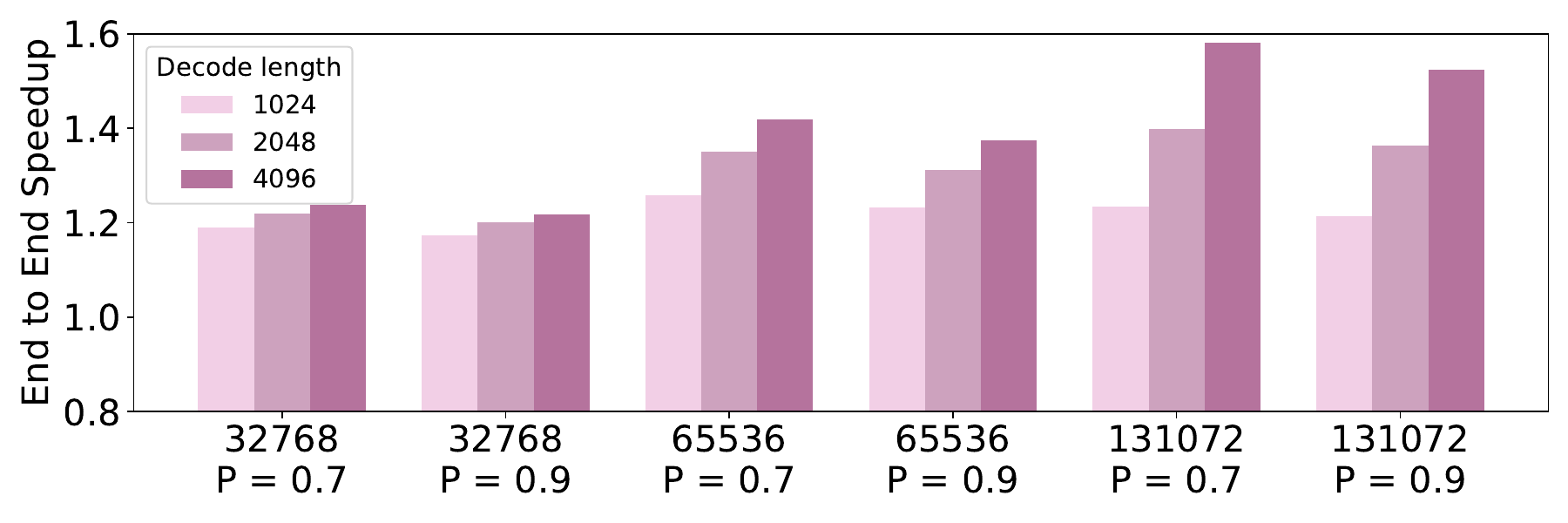}
    \caption{End-to-end speedup of \sys compared to the full attention.}
    \label{fig:e2e_speedup}
\end{figure}

\textbf{Latency of attention during decode.}
In the decode stage, \sys identifies and performs attention on critical tokens. We break down this process into four parts: 1) Cluster sorting, where the clusters are ranked based on the dot product of centroids and queries, 2) Distribution fitting, where \sys samples a small portion of tokens and derives the attention score to identify the token budget for each attention head, 3) performing attention for the selected tokens. \Cref{fig:decoding_time} shows the latency of this breakdown for different sequence lengths. 

The latency of sparse attention during decode is reduced significantly, while the overheads of sorting and distribution fitting remain low across various sequence lengths. Overall, \sys achieves up to \attnspeedup speedup compared to the full attention.

\subsubsection{Ablation Study for Query Head Union}

We evaluate the benefits of taking the union of grouped query heads versus computing attention for each query head individually. As shown in \fig{fig:union} Across different context lengths and ratio $P$, taking unions can achieve up to $1.65\times$ attention speedup, due to the reduced memory loading.

\subsubsection{End-to-end performance}
\label{sec:eval:e2e}

We compute the end-to-end performance of \sys with different output tokens, sequence lengths, and ratios in \Cref{fig:e2e_speedup} considering the prefill stages and the clustering overhead. Overall, \sys achieves a speedup up to \etoespeedup compared to full attention.

\section{Conclusion}

We presented \sys, a sparsity-adaptive attention mechanism for efficient long-context LLM inference. Unlike fixed token budget methods, \sys dynamically selects tokens based on cumulative attention scores, adapting to variations in attention sparsity. By leveraging clustering-based sorting and distribution fitting, \sys accurately estimates token importance with low overhead. Our results showed that \sys outperforms existing sparse attention methods, achieving higher accuracy and significant inference speedups, making it a practical solution for long-context LLMs.

\bibliography{_references}
\bibliographystyle{icml2025}
\newpage
\appendix
\section{Bound of approximation error}
\label{sec:appendix-bound-proof}

In \Cref{subsec:analysis-attn-approx}, the upper bound of $\epsilon(I)$ can be derived as
\begin{align}
    \epsilon(I)=&\|o-\tilde{o}(I)\| \\
    =&\|o-\frac{1}{p(I)}\sum_{i\in I}s_iv_i\|\\
    =&\|\sum_{i\in I}s_iv_i+\sum_{i\notin I}s_iv_i-\frac{1}{p(I)}\sum_{i\in I}s_iv_i\|\\
    =&\|\left(1-\frac{1}{p(I)}\right)\sum_{i\in I}s_iv_i+\sum_{i\notin I}s_iv_i\|\\
    \leq&\|\left(1-\frac{1}{p(I)}\right)\sum_{i\in I}s_iv_i\|+ \|\sum_{i\notin I}s_iv_i\|\\
    \leq&\left|\left(1-\frac{1}{p(I)}\right)\right|\sum_{i\in I}s_i\|v_i\| + \sum_{i\notin I}s_i\|v_i\|\\
    \leq&\left(\frac{1}{p(I)}-1\right)p(I)\max_i\|v_i\|+(1-p(I))\max_i\|v_i\| .
\end{align}
Hence we can get
\begin{equation}
    \epsilon(I)\leq 2(1-p(I))\max_i\|v_i\| .
\end{equation}
Both (10) to (11) and (11) to (12) are based on triangle inequality.

\section{Algorithm of \DF}
\begin{algorithm}[h]
\caption{Estimating Token Budget via Distribution Fitting}
\label{alg:token_selection}
\begin{algorithmic}[1]
\STATE \textbf{Input:} Token sequence unpacking from sorted clusters $\{x_1, x_2, \dots, x_n\}$, query $Q$, weight percentage threshold $P$, initial token count $N$, head dimension $d$
\STATE \textbf{Output:} Token budget $K$
\STATE
\STATE Compute $\mu_1$ and $\mu_2$ as the means of $\exp(x_i \cdot Q/ \sqrt{d})$ within fixed windows around $p_1$ and $p_2$. Solve for parameters $a$ and $b$ in $y = a/x + b$ the two data points.
\STATE
\STATE Initialize array $w_i$ to store the simulated attention scores for all tokens
\FOR{$i = 1$ to $n$} 
    \STATE \quad If $i \leq N$, $w_i = \exp(x_i \cdot Q/ \sqrt{d})$
    \STATE \quad Else, $w_i = a/i + b$
\ENDFOR
\STATE Compute the minimal $k$ such that the cumulative sum $\sum_{1}^k w_i \geq P \cdot \sum^{n}_{1}{w_i}$.
\STATE
\STATE \textbf{return} $k$
\end{algorithmic}
\end{algorithm}

\end{document}